# La traduction des noms propres : une étude en corpus

Emeline LECUIT[1], Denis MAUREL[2], Duško VITAS[3]
Université François Rabelais Tours, [1]LLL, [2]LI, France
[3]Université de Belgrade, Serbie

**Résumé** : Dans cet article, nous abordons le problème de la traduction des noms propres. Nous présentons notre hypothèse, selon laquelle la thèse très répandue de la non-traductibilité des noms propres peut être contredite. Puis, nous décrivons la construction du corpus multilingue aligné que nous utilisons pour illustrer notre propos. Nous évaluons enfin les apports et les limites de ce corpus dans le cadre de notre étude.

**Abstract** : In this paper, we tackle the problem of the translation of proper names. We introduce our hypothesis according to which proper names can be translated more often than most people seem to think. Then, we describe the construction of a parallel multilingual corpus used to illustrate our point. We eventually evaluate both the advantages and limits of this corpus in our study.

**Mots clés** : noms propres, traduction, corpus multilingue aligné

## Introduction

Les noms propres se traduisent-ils ? Nous tentons dans cet article de nuancer l'affirmation très répandue selon laquelle les noms propres ne se traduisent pas. Nous partons de l'hypothèse selon laquelle les noms propres sont, comme toute autre unité linguistique, susceptibles de subir des modifications lors de leur passage d'un texte en langue-source à un texte en langue-cible. Nous proposons donc de présenter la constitution d'un corpus parallèle aligné pour l'étude de la traduction des noms propres. Dans une première partie, nous introduirons la problématique et l'objectif de notre travail. Dans une deuxième et une troisième parties nous présenterons notre corpus et la méthodologie utilisée pour mener à bien notre étude. Puis, dans une quatrième partie, nous exposerons nos résultats. Nous aborderons les apports et les limites de notre corpus dans une cinquième partie, avant notre conclusion.

## 1. Problématique : la traduction des noms propres

Dans l'esprit de beaucoup, les noms propres ne se traduisent pas. « Je m'appelle Jean et si je vais à New York, je m'appelle toujours Jean », voilà une des réponses typiques à l'énonciation d'une possible traductibilité des noms propres. La plupart des grammaires présentent même la non-traductibilité des noms propres comme l'une des règles définitoires de la catégorie des noms propres, au même titre que d'autres règles (facilement mises à mal elles aussi) d'ordre syntaxique, typographique ou autre[1]. Les logiciens et linguistes ne sont pas non plus innocents à la propagation de cette idée reçue. Comme défenseur de cette théorie, nous pouvons citer George Moore[2], pour qui « Tous les noms propres […] doivent être rigoureusement respectés », ou encore Georges Kleiber (1981) pour qui, suivant sa théorie du nom propre comme prédicat de dénomination, « toute modification aboutit, non à une traduction d'un nom propre, mais à un nouveau nom propre ».

Récemment certaines études ont cependant ouvert de nouvelles perspectives concernant la traduction du nom propre. D'absolument intraduisible, le nom propre est d'abord devenu traduisible pour certaines exceptions (Algeo, 1973 ou Delisle, 1993), avant de devenir pour certains une unité de traduction à part entière : Ballard (2001) et Grass (2002), notamment, ont proposé d'étudier la traduction des noms propres, le premier en anglais, le second en allemand.

Notre étude, dans la lignée des deux dernières citées, mais élargie à un corpus dans dix langues différentes, montre que les noms propres, selon leur type, selon leur usage ou encore selon la langue cible de la traduction, sont sujets à tous les procédés de traduction existants, du report simple à la traduction enrichie en passant par le calque, la modulation, l'équivalence, etc. (Agafonov, 2006). Nous présentons ici un travail qui suit une démarche « *corpus-based* » ou exploitant un corpus (Tognini-Bonelli, 2001). En effet, à l'inverse d'une démarche « *corpus-driven* » ou émanant d'un corpus, qui voit l'élaboration d'un corpus en vue d'y repérer et d'en extraire des phénomènes linguistiques remarquables sans hypothèses préconçues, ni idées *a priori*, notre corpus se veut un inventaire de faits de langue précieux et approprié pour soutenir notre hypothèse et retirer des exemples précis pour l'illustrer. Il agit ici comme un matériel de support.

---

[1] Les critères définitoires de la catégorie des noms propres les plus souvent relevés sont : la majuscule, l'absence d'article, l'incompatibilité avec les déterminants, la mono-référentialité, l'absence de sens, ou encore, donc, l'intraduisibilité de ces derniers.

[2] Cité dans Ballard, 2001.

## 2. Présentation du corpus

Nous utilisons, pour étudier le nom propre en traduction, un corpus multilingue et aligné. Pour cette étude, nous avons choisi le roman de Jules Verne, *Le Tour du monde en quatre-vingts jours*. Ce choix a été motivé par deux raisons principales : l'existence de traductions dans presque toutes les langues européennes et leur disponibilité, ainsi que la présence d'un nombre non négligeable de noms propres tout au long du récit.

En effet, le roman de Verne, de par sa grande popularité, s'est retrouvé traduit dans de très nombreuses langues. Il s'agit même du roman de Jules Verne ayant été le plus traduit. Les années passant et les technologies évoluant, ce texte est désormais disponible dans de très nombreuses langues et de façon le plus souvent libre en format numérique.

D'autre part, en parcourant le roman, on s'aperçoit rapidement de l'abondance des noms propres de toutes sortes au fil des chapitres. Cette présence s'explique bien sûr par l'intrigue développée dans le récit. Il est peut-être utile de rappeler ici que *Le Tour du Monde en quatre-vingts jours* suit le gentleman Phileas Fogg et son valet Passepartout pour un tour du monde express que le héros se doit d'achever en moins de quatre-vingts jours s'il veut remporter son pari et ne pas perdre toute sa fortune. On y retrouve donc des noms propres appartenant à presque tous les types, aussi bien réels que fictifs (ce qui nous autorisera une réflexion sur la traduction de ces derniers). Pour rappel, les noms propres peuvent être classés en quatre grands types, qui sont les anthroponymes (noms de personnes individuels ou collectifs, de type ethnonymes, noms d'organisation, etc. mais aussi noms d'animaux), les toponymes (noms de lieux au sens large), les ergonymes (noms d'objets ou de produits de fabrication humaine) et les pragmonymes (noms d'événements), chacun de ces types pouvant être divisé en sous-types. Bien que, pour beaucoup, la plupart des noms propres se présentent sous la forme de ce que Jonasson (1994) a nommé les noms propres « purs » (l'élément ou les éléments constitutif(s) de ces noms propres étant emprunté(s) à un stock de noms ne pouvant être utilisés que comme noms propres), il ne faut pas oublier que les noms propres peuvent aussi être des noms propres « mixtes » ou « à base descriptive » (*ibid.*) (à savoir constitués d'un mélange de noms propres et d'éléments empruntés au lexique commun, noms et adjectifs la plupart du temps, ou uniquement d'éléments empruntés au lexique commun). Ainsi, *Phileas Fogg* est bien un nom propre, tout comme *l'Institution Royale de la Grande-Bretagne*, ou encore *la mer Rouge*.

Le texte, pour sa version française, comporte 3 415 noms propres (dont 519 différents). Ces noms propres représentent 8,6% des caractères du texte et 8% des mots du texte[3]. Ce pourcentage de noms propres et leur diversité fait du texte de Jules Verne un candidat idéal pour notre étude.

Nous avons donc constitué un corpus composé de différente version, c'est-à-dire dans des langues différentes, du roman de Jules Verne. Ce corpus a été aligné, autrement dit les phrases traduites ont été reliées au niveau de segments équivalents, ce qui nous permet une comparaison facile d'une langue à l'autre. Pour mener à bien notre étude, nous avons eu besoin d'utiliser certains outils de traitement automatique des langues que nous présentons dans la section suivante.

---

[3] Cette couverture d'un texte par les noms propres atteint à peu près 10 % dans les textes journalistiques, comme le souligne Coates-Stephens (1993).

*La traduction des noms propres : une étude en corpus*

## 3. Présentation des outils

Dans un premier temps, onze versions du roman de Verne ont été collectées[4] : en français, anglais (deux versions différentes), allemand, espagnol, portugais, italien, serbe (en alphabet latin[5]), polonais, bulgare et grec. Cette sélection, limitée par la disponibilité respective des différentes versions, est assez représentative du paysage linguistique européen. Nous sommes, en effet, en présence ici de langues appartenant aux différentes familles de langues issues de l'Indo-Européen présentes en Europe : quatre langues latines, trois langues slaves, deux langues germaniques et le grec. Pour être tout à fait représentatif des familles de langues indo-européennes parlées en Europe, il aurait été nécessaire d'inclure dans notre corpus une langue celte (l'irlandais, par exemple) ainsi qu'une version du texte en balte. Malheureusement, ces versions du texte, même si elles existent, ne sont pas disponibles en accès libre. Nous « limitons » donc notre corpus à une sélection de dix langues.

      Le roman a été traduit de nombreuses fois en anglais. Nous proposons, dans notre corpus, deux versions anglaises différentes. La première version, rédigée en hâte à la vue du succès du roman en France, a rapidement été critiquée pour sa médiocrité. Pire encore, Jules Verne lui-même a fait l'objet de critiques, étant qualifié par certains d'auteur pour enfant. Cette version restant néanmoins la plus éditée à ce jour[6], il est normal qu'elle trouve sa place dans notre corpus. La deuxième version, plus récente, que nous proposons d'étudier respecte davantage le texte original (en matière de contenu, mais aussi de forme, ce que la première version semblait avoir ignoré). Elle a été jugée « *by far the best translations/critical editions available* » par le magazine Science-Fiction Studies. L'étude parallèle de ces deux versions nous permet d'étudier un autre facteur influant sur la traduction des noms propres, qui est la fidélité de la traduction.

---

4 Il s'agit de versions du roman en accès libre sur le net.
5 Le serbe utilise indifféremment l'alphabet cyrillique ou latin.
6 C'est cette même version qui est aujourd'hui éditée chez Penguin.

Les différentes versions du texte ayant été prélevées, nous nous retournons vers la version originale. Le texte français subit alors une étape fondamentale : le repérage et l'étiquetage des noms propres qu'il contient. Nous avons réalisé l'extraction des noms propres dans le texte en version originale grâce à l'outil CasSys (Friburger, 2004), disponible sur la plateforme de traitement linguistique Unitex[7] (Paumier, 2006), et aux transducteurs regroupés sous le nom de CasEN[8], développés pour la campagne d'évaluation des Systèmes de Transcription Enrichie d'Emissions Radiophoniques (ESTER)[9]. Après une phase de pré-analyse (division du texte en phrases, étiquetage avec les dictionnaires, etc.), l'application CasSys peut être lancée. Une série de transducteurs à nombre fini d'états est appliquée au texte. Chaque transducteur contient une grammaire locale décrivant les différents contextes pouvant indiquer la présence d'un objet à repérer. Dans notre cas, les transducteurs sélectionnés permettent la localisation des différents noms propres. Des balises, indiquant le type du nom propre localisé, sont automatiquement introduites dans le texte. Un aperçu de ce balisage est visible ci-dessous (Figure 1). Il s'agit de la première phrase du roman après balisage.

> En l'année 1872, la maison portant le numéro 7 de <ENT type="loc.line">Saville-row</ENT>, <ENT type="loc.line">Burlington Gardens</ENT> -- maison dans laquelle <ENT type="pers.hum">Sheridan</ENT> mourut en 1814 --, était habitée par <ENT type="pers.hum">Phileas Fogg, esq.</ENT>, l'un des membres les plus singuliers et les plus remarqués du <ENT type="org">Reform-Club de Londres</ENT>, bien qu'il semblât prendre à tâche de ne rien faire qui pût attirer l'attention.

**Figure 1: Étiquetage du texte avec CasSys (extrait)**

Une fois les textes rassemblés et le texte français étiqueté, la troisième étape consiste à aligner les traductions avec le texte français. Notre texte français est aligné avec les autres versions du texte au niveau des segments équivalents, c'est-à-dire au niveau des segments ayant le même contenu sémantique (mais pas forcément la même taille, ce qui est une des difficultés (voir Vitas, 2008)). Pour mener à bien cette étape, nous faisons appel à l'outil Xalign (Loria, 2006), qui fait aussi partie des outils disponibles sous Unitex. L'alignement se fait entre deux textes à la fois, l'outil Xalign n'autorisant le travail que sur des bitextes.

---

[7] http://www-igm.univ-mlv.fr/~unitex/

[8] Cette cascade dédiée à la recherche des entités nommées est disponible, sous licence LGPL-LR, à l'URL : http://tln.li.univ-tours.fr/Tln_CasEN.html

[9] http://www.afcp-parole.org/camp_eval_systemes_transcription/

   Avant de procéder à l'alignement, chaque version du texte est transformée en un format TEI. Chaque phrase, paragraphe, et division du texte est encadré par des balises, respectivement <s>, <p> et <d> et reçoit un code d'identification (par exemple, « d1p1s1 » correspond à la première phrase du premier paragraphe de la première division du texte)[10]. Tous ces marqueurs fonctionnent comme des points d'ancrage explicites qui facilitent l'alignement. D'autres points d'ancrage potentiels, comme les cognats[11] par exemple, participent également à l'alignement des textes. L'aligneur extrait le chemin complet optimum et nous obtenons des équivalences 1:1, 1:2 ou encore 2:1 entre la version française et la version alignée. Grâce à la fenêtre d'alignement générée, il est possible de visualiser les équivalents de traduction de manière claire, une phrase du texte français étant reliée visuellement par une ligne à son équivalent dans la version étrangère (voire Figure 2). Des corrections manuelles de l'alignement sont parfois nécessaires, une vérification systématique a donc été effectuée pour chaque bitexte aligné produit.

   Le fait qu'Xalign fonctionne sous Unitex nous permet de profiter des avantages offerts par ce logiciel de traitement de corpus. Unitex permet en effet différentes actions sur les textes, telles que des requêtes linguistiques complexes sous la forme de graphes. Les bitextes ayant été créés, nous les réunissons sous la forme d'un grand tableau permettant l'observation des phénomènes sur toutes les langues en même temps. Nous créons ainsi un multitexte réunissant onze versions différentes d'un même texte[12]. Un extrait de ce multitexte (la première phrase du roman dans toutes les versions) est présenté Figure 3, ainsi qu'un gros plan sur cet extrait (voir la Figure 4, ci-dessous).

---

[10] Dans notre cas, le terme « division » correspond à un chapitre du livre.

[11] Les cognats sont les chaînes invariantes d'un texte à l'autre.

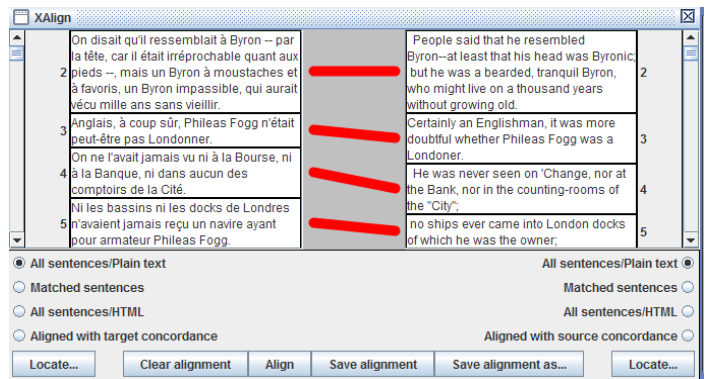

**Figure 2: Alignement d'un bitexte (Xalign sous Unitex)**

## 4. Résultats

La disponibilité de ce corpus multilingue nous permet d'appréhender le problème de la traduction des noms propres de manière très large en ce qui concerne le nombre de langues alignées et les différentes catégories de noms propres représentées. Les nombreux outils de TAL à notre disposition pour l'alignement des textes, l'étiquetage des noms propres, le transfert d'information ou la localisation d'éléments par requêtes complexes, nous permettent de voir que :
- selon leur genre (noms propres de fiction ou noms propres réels),
- selon leur type (anthroponymes, toponymes, dérivés ou encore référents culturels divers),

---

12 Ce qui est en fait très rare. Les seuls multitextes conséquents disponibles à ce jour étant pour la plupart les textes de lois de l'Union Européenne.

*La traduction des noms propres : une étude en corpus*

**Figure 3: Corpus multilingue**
**(Alignement de la première phrase dans les onze versions)**

| FRA-NP | BUL | ENG1 |
|---|---|---|
| En l'année 1872, la maison portant le numéro 7 de {ENT type'"loc.line"}Saville-row{/ENT}, {ENT type'"loc.line"}Burlington Gardens{/ENT} -- maison dans laquelle {ENT type'"pers.hum"}Sheridan{/ENT} mourut en 1814 --, était habitée par {ENT type'"pers.hum"}Phileas Fogg, esq.{/ENT}, l'un des membres les plus singuliers et les plus remarqués du {ENT type'"org"}Reform-Club de Londres{/ENT}, bien qu'il semblât prendre à tâche de ne rien faire qui pût attirer l'attention | През 1872 година в къщата на "Савил роу" № 7, Бърлингтън Гардънс – същата, в която през 1814 година почина Шеридан, – сега живееше Филиас Фог. ой беше един от най-странните и видни членове на Реформаторския лондонски клуб, въпреки че сякаш се стараеше да | Mr. Phileas Fogg lived, in 1872, at No. 7, Saville Row, Burlington Gardens, the house in which Sheridan died in 1814.He was one of the most noticeable members of the Reform Club, though he seemed always to avoid attracting attention; |

**Figure 4 : Corpus multilingue (Alignement de la première phrase dans trois versions)**

- selon leur usage (comme simple signifiant, mais aussi dans leur usage modalisé ou rhétorique),
- selon leur construction (noms propres « purs », noms propres « modifiés », etc. (Jonasson, 1994)),
- selon la langue cible de la traduction (différents comportements morphologiques, différents alphabets, etc.),

les noms propres subissent, lors de leur passage d'une langue à une autre, bon nombre de procédés de traduction existants.

*La traduction des noms propres : une étude en corpus*

Le texte original comporte 3 415 noms propres (519 différents) répartis comme suit (voir Figure 5) :

| hypertypes | nombre total d'occurrences | nombre d'occurrences différentes |
|---|---|---|
| anthroponymes | 2079 | 162 |
| toponymes | 1142 | 320 |
| ergonymes | 186 | 31 |
| pragmonymes | 8 | 6 |

**Figure 5 : Les noms propres dans le texte original**

Notre étude n'étant pas encore complètement achevée, nous présentons ici quelques-uns des phénomènes observés pour 10 % des noms propres de chaque type. Pour ce faire, nous avons prélevé les trente-deux toponymes, les seize anthroponymes, les trois ergonymes et le pragmonyme le plus souvent répétés dans le texte original, et observé leur comportement lors de leur passage dans les différentes langues-cibles. Ces résultats sont présentés dans la Figure 6. Ces cinquante-deux noms propres, tous différents, représentent à eux seuls 2 029 occurrences dans le texte de départ, soit près de 60 % des noms propres apparaissant effectivement dans le texte[13].

| Langue-cible | emprunt | assimilation | calque partiel ou total | Absence de traduction | Autres procédés |
|---|---|---|---|---|---|
| **Anglais (1ère version)** | 69,1% | 11,3% | 2,2% | 12,1% | 5,4% |
| **Anglais (2nde version)** | 74,2% | 13,2% | 2,0% | 6,6% | 4,1% |
| **Allemand** | 79,7% | 10,6% | 3,7% | 5,2% | 0,6% |
| **Polonais** | 31,1% | 53,4% | 4,5% | 10,7% | 0,2% |
| **Serbe (latin)** | 4,9% | 89,1% | 4,5% | 0,3% | 1,3% |
| **Bulgare** | 0,0% | 90,6% | 6,3% | 2,5% | 0,7% |
| **Grec** | 0,0% | 86,8% | 4,6% | 3,5% | 5,1% |
| **Italien** | 72,0% | 21,5% | 2,6% | 3,0% | 1,0% |
| **Portugais** | 73,7% | 16,0% | 5,9% | 4,1% | 0,2% |
| **Espagnol** | 51,6% | 15,5% | 25,1% | 7,4% | 0,4% |

**Figure 6 : Procédés de traduction (résultats)**

---

[13] Les noms propres utilisés à l'intérieur d'autres noms propres mixtes, n'ont pas été comptés ici. Ainsi l'occurrence *Londres* dans *Insitution de Londres* n'a pas été comptabilisée parmi les occurrences de *Londres*.

On considère, en traduction, l'emprunt[14] ou le report, comme le degré zéro de la traduction. Ce phénomène, le plus fréquent dans la majorité des langues de notre corpus, est observé pour tous les anthroponymes du texte lors de leur passage du texte français aux textes anglais, espagnol, italien, portugais et polonais.

Ainsi on retrouve *Phileas Fogg* dans toutes ces versions du texte (avec cependant des phénomènes d'assimilation, notamment en polonais où *Phileas* devient *Fileas*). Une exception cependant à cette utilisation systématique du report pour les anthroponymes : l'auteur de la version espagnole a choisi de traduire le nom fictif *Passepartout*[15], en utilisant une traduction littérale totale. Il a donc renommé *Jean Passepartout*, *Juan Picaporte* (« picaporte » qui signifie littéralement « passepartout »), ce qui explique le pourcentage important de « calque » dans la colonne espagnole du tableau.

Pour ce qui est des versions en langues grecque ou bulgare, des phénomènes de transcription sont observables, ce qui est attendu de versions utilisant un alphabet différent (en l'occurrence l'alphabet grec et l'alphabet cyrillique respectivement). On a, par exemple *Aouda* en français, *Αούντα* en grec et *Ауда* en bulgare. Ces phénomènes ont été comptabilisés dans la colonne « assimilation ». On peut donc dire qu'il n'y a pas d'emprunts dans ces langues.

Plus curieusement, ce même phénomène de transcription se retrouve aussi dans la version serbe où tous les noms propres subissent des phénomènes d'adaptation

---

[14] Pour le polonais, nous avons différencié, dans notre étude, les noms propres empruntés mais non déclinés et les noms propres empruntés mais déclinés normalement, à savoir respectivement 26,5% et 4,6% des phénomènes observés. C'est le pourcentage global d'emprunts qui est donné ici.

15 La liberté de l'auteur d'ici traduire le nom de *Passepartout* s'explique du fait du caractère fictif de ce nom. Il est à noter que, de tous les noms de personnes (ou anthroponymes) utilisés par Verne, seul *Passepartout* semble avoir été emprunté au lexique commun. Tous les autres ont, eux, été puisés dans ce qui pourrait être appelé la classe des noms propres existants, que ce soient les prénoms ou les noms de famille.

*La traduction des noms propres : une étude en corpus*

phonétique ou graphique, ce qui permet l'adaptation de ceux-ci aux différentes exigences orthographiques du serbe tout en respectant la prononciation locale des noms propres (*Fix* devient donc *Fiks* en serbe).

Remarquons que l'assimilation est aussi présente dans d'autres langues pour certains noms propres. C'est même pour la plupart de ces langues, le deuxième procédé de traduction des noms propres (en pourcentage d'utilisation). Par exemple, *Aouda*, devient *Aouida* en espagnol et *Auda* en Italien.

D'autres phénomènes sont observés, notamment sur le jeu des articles, qui change d'une langue à l'autre. Nous avons dans notre corpus des langues sans articles (comme le polonais, par exemple), mais aussi des langues dans lesquelles les articles ne sont pas nécessairement employés dans les mêmes contextes (on pense au jeu des articles avec les noms de pays en français qui n'est pas reproduit dans les autres langues du corpus). Des changements de genre et de nombre peuvent également survenir d'une langue à l'autre, comme l'*Inde* (au singulier en français) qui devient *Indie* en polonais, un nom pluriel.

Des différences de construction morphologique apparaissent également. On note une postposition de l'article en bulgare. On relève également le jeu des déclinaisons qui, selon les langues, peut donner à partir d'un même nom propre en français jusqu'à sept formes dans la langue cible[16].

On trouve dans la version polonaise la création d'un paradigme flexionnel spécifique aux noms propres apparaissant comme étrangers. Ainsi, les noms propres se terminant en un *e* muet ou avec le son *i* par exemple se verront dotés d'une apostrophe (autrement jamais utilisée en polonais) entre la fin du nom et sa déclinaison. On trouve ainsi *Cromarty'ego* ou encore *Mascarille'a*. D'autres noms propres étrangers, comme *Passepartout* par exemple, ne sont ni traduits ni fléchis.

---

[16] Deux cas pour le bulgare, quatre pour l'allemand, cinq pour le grec et sept pour le polonais et le serbe.

Pour ce qui est des phénomènes de « calque partiel ou total » référencés dans le tableau, ils concernent, dans toutes les langues, la majorité des noms propres que nous avons qualifiés plus haut de noms propres « mixtes » ou « à base descriptive ».

On peut également relever dans quelques langues (anglais, polonais et serbe) des exemples de traduction par une forme dérivée (création d'un adjectif à partir du nom propre). On trouve ainsi dans la première version anglaise la forme *Byronic* (en référence à *Byron*). La version en serbe, langue qui possède la caractéristique de pouvoir former des adjectifs possessifs à partir de noms propres en ajoutant le suffixe *–ov* ou *-ev* pour les noms masculins ou le suffixe *–in* pour les noms féminins, nous propose la dérivation *Paspartuov* à partir du nom propre *Passepartout* et la dérivation *Audin* à partir du nom propre *Auda* (avec effacement de la voyelle finale).

Ces deux phénomènes (flexion et dérivation) peuvent se combiner et générer ainsi un certain nombre de formes pour un même nom propre. C'est le cas notamment en serbe, où l'on trouve neuf formes de Passepartout, quatre fléchies à partir du lemme *Paspartu* en tant que nom, à savoir *Paspartu* (nominatif), *Paspartua* (accusatif ou génitif), *Paspartuu* (datif), *Paspartuom* (instrumental) et cinq à partir du lemme *Paspartuov* en tant qu'adjectif possessif, à savoir *Paspartuov* (nominatif), *Paspartuova* (génitif et accusatif), *Paspartuovu* (datif), *Paspartuovih* (génitif pluriel) *Paspartuovim* (instrumental).

Les « noms propres modifiés », subissent également le passage de la langue source aux différentes langues cibles. Le démonstratif dans l'expression *« Ce Phileas Fogg était-il riche ? »* disparait dans la première version anglaise ainsi que dans les versions portugaise, polonaise et espagnole, par exemple. Dans la version espagnole, le groupe proprial *« L'intraitable Fogg »* est, quant à lui, divisé et l'adjectif est postposé et étendu en un groupe adjectival complexe. On peut donc lire « *[...]Fogg, quien, tan intratable y tan abotonado como siempre,[...]* ».

Les « absences de traduction », notamment dans la première version anglaise et la version polonaise, touchent pour la plupart les anthroponymes. Dans la plupart des cas, ils ont été remplacés par des anaphores pronominales, mais aussi parfois par des descriptions définies.

Les « autres procédés » sont variés, des réductions, très nombreuses en anglais (dans la première, comme dans la deuxième version), aux dérivations (changement de catégorie lexicale, la plupart du temps le nom propre devient adjectif, comme en serbe, voir plus haut) en passant par l'utilisation de désignations distinctes (par choix, comme par exemple dans la deuxième version anglaise où le traducteur préfère « *Britain* » à « *England* », ou par erreur, comme quand le traducteur serbe utilise *Endrju Tomas* à la place d'*Endrju Stjuart*[17]).

Une manifestation d'antonomase du nom propre a également attiré notre attention. L'antonomase du nom propre intervient lorsqu'un nom propre est utilisé pour signifier un nom commun. On trouve dans le texte français *« un de ces Frontins ou Mascarilles »*. Le statut particulier de l'antonomase, intermédiaire entre le nom commun et le nom propre, place le traducteur face à un choix. Il peut soit décider de respecter l'antonomase et donc de réutiliser un nom propre dans sa version : c'est ce que propose presque tous les traducteurs des versions étudiées (on trouve par exemple dans la seconde version anglaise *« one of those Frontins or Mascarilles »*). Il peut, sinon, décider de passer par l'utilisation d'un nom commun, ce que propose le traducteur de la première version anglaise : *« one of those pert dunces depicted by Molière »*.

---

[17] Nous savons qu'il s'agit d'une erreur car, les autres fois c'est bien *Endrju Stjuart* qui est utilisé.

Dans la lignée du phénomène précédent[18], on peut évoquer le terme *« linge en toile de Saxe »*, qui devient *« the finest linen »* dans la première version anglaise et *« Saxony table-cloth »* ou d'autres termes reprenant le nom propre *Saxe*, dans la seconde version anglaise et dans les autres langues.

Tous les exemples que nous venons d'évoquer sont rendus facilement repérables et observables grâce à l'alignement des textes et au repérage préalable des éléments nous intéressant dans le texte français. Il apparait évident que de nombreux exemples contredisent la théorie de la non-traductibilité des noms propres.

## 5. Les apports et les limites de notre corpus

Nous discutons dans cette partie les avantages et inconvénients offerts par notre corpus.

Nous présentons ici un multitexte en dix langues, comprenant onze versions d'un roman de la fin du XIXème siècle. Les multitextes proposant autant de langues sont plus que rares et proposent habituellement des textes de lois (cf. par exemple, les textes de lois de l'Union Européenne). C'est donc un tout nouveau genre de corpus qui est introduit ici et permet un travail simultané et croisé sur un phénomène linguistique dans une dizaine de langues européennes.

Le roman est certes un peu ancien, mais cela nous permet d'échapper à la contrainte onéreuse des droits d'auteur.

Autre avantage, notre corpus reste extensible. La disponibilité des textes est, comme nous l'avons souligné, relativement large. Nous n'avons certes pas trouvé de version finnoise du texte, mais nous avons à notre disposition notamment les versions arabe, hongroise et chinoise du texte. Notre étude s'arrêtant aux langues indo-européennes, nous avons décidé de ne pas inclure ces versions à notre corpus, mais leur étude pourra faire l'objet d'un travail ultérieur.

---

[18] L'antonomase du nom propre, tout comme le terme de cet exemple ou encore les phénomènes des expressions idiomatiques comprenant des noms propres, sont trois cas dans lesquels le nom propre a perdu sa signification (voir Tran et Maurel, 2006).

Nous insistons sur le fait que le roman de Jules Verne est plus que propice à l'étude des noms propres. En effet, nous rappelons que leur nombre est conséquent, mais que, surtout, ils couvrent toutes les catégories de noms propres figurant dans notre typologie. On pourra regretter que certains phénomènes soient peu représentés quantitativement. On pensera par exemple aux noms d'animaux (un seul référent dans le texte) ou aux noms d'évènements (six références seulement). Mais il n'est pas certain que d'autres genres littéraires auraient offert un éventail aussi large de noms propres.

On pourra justement nous reprocher le genre restrictif du corpus. Le roman littéraire n'est qu'un genre parmi d'autres et une étude linguistique doit couvrir plusieurs genres textuels pour pouvoir affirmer des hypothèses. Notre étude reste donc à compléter sans doute par une étude sur un autre type de corpus tel qu'un corpus journalistique, juridique, etc. On pourrait envisager l'étude sur des flux d'actualités.

**Conclusion**

L'hypothèse largement répandue de l'intraduisibilité des noms propres est souvent contredite grâce à l'étude de notre corpus créé spécialement pour observer ce phénomène. Certes, les emprunts sont nombreux, mais d'autres phénomènes n'en restent pas moins appréciables.

Notre démarche *corpus-based* nous permet d'analyser les tendances en matière de traduction des noms propres et d'illustrer notre point de vue. Une étude détaillée (à suivre) de tous les exemples de noms propres présents dans notre corpus nous permettra de produire un travail de description formelle et détaillée des caractéristiques morphologiques des noms propres en langue-source et dans les langues-cibles. Ces indications pourront fournir une aide précieuse au traducteur quand celui-ci sera confronté à des difficultés lors de la traduction de noms propres.

Notre étude pourra être étendue à d'autres langues ainsi qu'à d'autres types de corpus même si la disponibilité d'un corpus dans les langues étudiées et comprenant le phénomène à observer (à savoir les noms propres), dans les mêmes mesures, aussi bien quantitative que qualitative, que notre corpus, parait peu envisageable.

**Références bibliographiques**

Agafonov C., Grass T., Maurel D., Rossi-Gensane N. et A. Savary (2006). « La traduction multilingue des noms propres dans PROLEX », in *Méta*, Vol.51, n°4 : 622-636.

Algeo J. (1973). *On defining the Proper Name*, Gainesville, Florida.

Ballard M. (2001). *Le nom propre en traduction*, Paris : Ophrys.

Bauer G. (1998). *Namenkunde des Deutschen*, Berlin, Germanistische Lehrbuchsammlung Band 21.

Coates-Stephens S. (1993). « The Analysis and Acquisition of Proper Names for the Understanding of Free Text », in *Computers and the Humanities*, Vol.26: 441-456.

Delisle J. (1993). *La traduction raisonnée. Manuel d'initiation à la traduction professionnelle de l'anglais vers le français*, Ottawa, Presses de l'Université d'Ottawa.

Friburger N. et D. Maurel (2004) «Finite-state transducer cascades to extract named entities in texts», in *Theoretical Computer Science*, Vol.313(1): 93-104.

Grass T. (2002). *Quoi ! Vous voulez traduire « Goethe » ? Essai sur la traduction des noms propres allemand-français*, Berne, Peter Lang.

Jonasson K. (1994). *Le nom propre : constructions et interprétations*, Louvain-la-Neuve : Duculot.

Kleiber G. (1981). *Problèmes de référence : descriptions définies et noms propres*, Paris, Klincksieck.

LORIA (2006), XAlign, http://led.loria.fr/outils/ALIGN.align.html

*La traduction des noms propres : une étude en corpus*

Paumier S. (2006). *Unitex 2.0 User Manual*, http://igm.univ-mlv.fr/~unitex~/

Tognini-Bonelli, E. (2001): *Corpus Linguistics at Work*, Amsterdam/Philadelphia: Benjamins.

Tran M. et D. Maurel (2006). « Prolexbase : Un dictionnaire relationnel multilingue de noms propres », in *Traitement automatique des langues*, Volume 47, n°3 : 115-139.

Vaxelaire J.-L. (2005). *Les noms propres – une analyse lexicologique et historique*, Champion.

Vinay J.-P. et J. Darbelnet (2004). «A methodology for Translation», in *The Translation Studies Reader*, Venuti L. (ed): Routledge :84-93.

Vitas D., Koeva S., Krstev C., et I. Obradovic (2008). « Tour du monde through the dictionaries », in *Proceedings of the 27th Conference on Lexis and Grammar*, L'Aquila :249-257.